\documentclass[sigconf,nonacm]{acmart}
\usepackage[table]{xcolor}
\definecolor{accbluebg}{HTML}{D6E8F7}   
\definecolor{accbluetext}{HTML}{0051A3} 
\definecolor{mibgray}{HTML}{E6E6E6}     
\usepackage{colortbl}
\AtBeginDocument{%
  }

\newcommand{\methodnameplain}{TRACE}
\newcommand{\methodname}{\textup{\texttt{\methodnameplain}}}
\usepackage{float}
\floatstyle{ruled}
\newfloat{algorithm}{tbp}{loa}
\floatname{algorithm}{Algorithm}
\newcounter{algline}
\newenvironment{pseudocode}{%
  \begin{list}{\arabic{algline}:}{\usecounter{algline}%
    \setlength{\leftmargin}{1.6em}\setlength{\labelwidth}{1.3em}%
    \setlength{\labelsep}{0.3em}\setlength{\itemsep}{0pt}%
    \setlength{\parsep}{0pt}\setlength{\topsep}{3pt}}%
}{\end{list}}
\newcommand{\pstep}[2][0]{\item\hspace*{#1em}#2}

\begin{document}

\title{\methodnameplain{}: Memory-Efficient Time Series Class-Incremental Learning via Temporal-Spectral Reconstruction}

\author[Junhui Mao]{\mbox{Junhui Mao\textsuperscript{1}}}
\author[Junru Zhang]{\mbox{Junru Zhang\textsuperscript{1}}}
\author[Liming Ye]{\mbox{Liming Ye\textsuperscript{1}}}
\author[Cheng Peng]{\mbox{Cheng Peng\textsuperscript{1}}}
\affiliation{%
  \institution{Zhejiang University}
  \city{Hangzhou}
  \country{China}
}

\author[Haoran Shi]{\mbox{Haoran Shi\textsuperscript{2}}}
\author[Xiaoli Tang]{\mbox{Xiaoli Tang\textsuperscript{2}}}
\affiliation{%
  \institution{Nanyang Technological University}
  \city{Singapore}
  \country{Singapore}
}

\author[Baokun Wang]{\mbox{Baokun Wang\textsuperscript{3}}}
\affiliation{%
  \institution{Ant Group}
  \country{China}
}

\author[Bin Dou]{\mbox{Bin Dou\textsuperscript{3}}}
\affiliation{%
  \institution{Ant Group}
  \country{China}
}

\author[Yabo Dong]{\mbox{Yabo Dong\textsuperscript{1}}}
\affiliation{%
  \institution{Zhejiang University}
  \city{Hangzhou}
  \country{China}
}

\author[Han Yu]{\mbox{Han Yu\textsuperscript{2}}}
\affiliation{%
  \institution{Nanyang Technological University}
  \city{Singapore}
  \country{Singapore}
}

\renewcommand{\shortauthors}{Junhui Mao et al.}
\makeatletter
\renewcommand{\@mkauthors@iii}{%
  \global\setbox\mktitle@bx=\vbox{%
    \unvbox\mktitle@bx\par\medskip
    \centering
    {\@authorfont\def\and{,\space}\authors\par}%
    \smallskip
    {\@affiliationfont
      \textsuperscript{1}Zhejiang University, Hangzhou, China\quad
      \textsuperscript{2}Nanyang Technological University, Singapore\par
      \textsuperscript{3}Ant Group, China\par}%
    \bigskip}}
\makeatother

\begin{abstract}
Web-connected sensing applications continuously generate multivariate time series with evolving behavioral categories, making Time Series Class-Incremental Learning (TSCIL) essential for adapting to new classes without forgetting previously acquired knowledge. However, preserving diverse and evolving class distributions under strict privacy and memory constraints remains a fundamental challenge. Existing approaches often face a trade-off between retaining sufficient historical knowledge for accurate recognition and minimizing the memory required for continual adaptation. To address these limitations, we propose \methodname{}, a memory-efficient, replay-free framework that preserves class-specific knowledge through generative modeling rather than sample rehearsal. \methodname{} employs a compact, frozen temporal-spectral representation and lightweight class-specific autoencoders to capture complex intra-class variations. By independently preserving each class model, \methodname{} avoids recursive generation and catastrophic forgetting, while enabling direct classification through reconstruction-based scoring without retaining historical samples or large statistical matrices. Extensive experiments across five TSCIL benchmarks demonstrate that \methodname{} achieves competitive classification accuracy while reducing mean retained memory by approximately 98.26\% compared with TS-ACL, the strongest evaluated analytical baseline, and achieving an approximately 33.43\% relative improvement in accuracy on GrabMyo.
\end{abstract}

\ccsdesc[500]{Computing methodologies~Online learning settings}
\ccsdesc[300]{Information systems~Data mining}

\keywords{Time Series Classification, Class-Incremental Learning, Continual Learning, Generative Classification, Reconstruction Scoring}

\maketitle
\hypersetup{pdfauthor={Junhui Mao, Junru Zhang, Liming Ye, Cheng Peng, Haoran Shi, Xiaoli Tang, Baokun Wang, Bin Dou, Yabo Dong, Han Yu}}

\section{Introduction}

The Web has become a critical infrastructure for acquiring, exchanging, and interpreting continuous sensor-generated data across distributed devices and services. In Web-connected applications, such as wearable activity monitoring and tele-rehabilitation, multichannel time series provide rich temporal information for extracting meaningful behavioral patterns from continuously evolving data streams~\cite{qiao2024class}. Unlike static data, these streams exhibit transient variations, periodic patterns, and multi-scale temporal dependencies~\cite{ismail2019deep,wu2022timesnet}. Moreover, as Web-connected applications continuously encounter new users, activities, and interaction contexts, the underlying behavioral categories evolve over time, making static classification models inadequate. This motivates Time Series Class-Incremental Learning (TSCIL)~\cite{qiao2024class}, which enables continuous acquisition of new behavioral knowledge while preserving previously learned classes. However, deploying TSCIL in distributed Web-based edge-cloud environments remains challenging, as privacy concerns restrict access to historical user data, while resource-constrained devices impose stringent limitations on memory consumption and synchronization overhead. Meanwhile, models should continuously recognize both previously learned and emerging classes without relying on task-identity information.

\begin{figure}[t]
  \centering
  \includegraphics[width=\columnwidth]{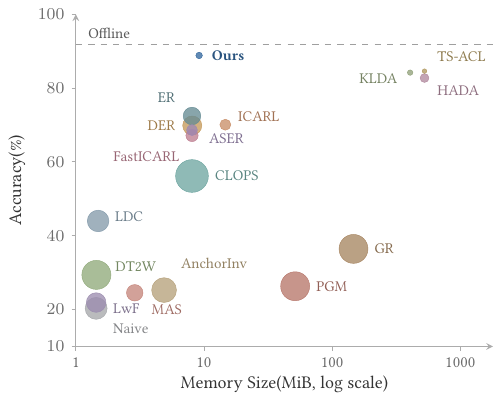}
  \caption{Mean final test accuracy versus retained memory across five datasets (25 runs per dataset). Bubble area represents the mean within-dataset variance across runs, and the dashed line indicates the offline joint-training reference.}
  \Description{Accuracy versus retained state for the same methods as Table 2, equally averaged across five datasets. Ours uses a frozen CNN with class-specific deterministic AEs. Its point is at 9.1654 MiB and 88.8538 percent. Circle area is 4.9 times the mean of five within-dataset sample run variances, without a size floor, and does not represent a confidence interval. The accuracy axis spans 10 to 100 percent. Offline is shown as a dashed horizontal reference.}
  \label{fig:fig1_tradeoff}
  \vspace{-0.9cm}
\end{figure}

Existing TSCIL paradigms struggle to seamlessly navigate these constraints, facing a stark trade-off between accuracy and memory (Figure~\ref{fig:fig1_tradeoff}). Regularization techniques~\cite{li2018lwf,aljundi2018mas} (e.g., LwF and MAS) suffer from catastrophic forgetting, while experience replay~\cite{rolnick2019experience,buzzega2020dark} violates privacy mandates. Generative replay~\cite{shin2017continual,qiao2024class}  (e.g., GR) bypasses data storage but introduces prohibitive synthesis overhead and cascading errors. Recently, analytic continual learning (ACL)~\cite{momeni2025klda,li2025tsacl,liu2026hada}  (e.g., KLDA, TS-ACL, and HADA)  addressed forgetting by freezing the backbone and dramatically expanding feature dimensions (e.g., $5{,}000$ to $8{,}000$). While mathematically robust, maintaining such massive covariance matrices demands hundreds of megabytes of persistent memory, rendering them intractable for resource-constrained edge architectures.

A pivotal bottleneck in memory-constrained TSCIL is the reliance on a \emph{globally shared parametric state}. Continuously updating shared boundaries to accommodate novel classes inevitably triggers representation drift and catastrophic forgetting, whereas indiscriminately expanding the model state triggers architectural bloat. This dilemma is particularly pronounced in time series, where disparate activities may share local waveform morphologies but diverge radically in frequency or periodic evolution~\cite{schafer2017fast,cui2016multi}. To resolve this, we ask a fundamental question: \emph{How can we support class expansion in time series recognition while preserving previously learned knowledge, without historical replay or high-dimensional feature expansion?}

In response, we introduce a novel replay-free TSCIL framework called \methodname{} (\textbf{T}emporal-Spectral \textbf{R}econstruction for \textbf{A}daptive \textbf{C}lass \textbf{E}xpansion). Our key insight is that continual recognition of evolving time series classes does not necessarily require repeatedly updating shared decision boundaries. Instead, a stable temporal-spectral representation can support independent modeling of class-specific variations. Guided by this insight, \methodname{} integrates complementary temporal dynamics and frequency characteristics into a compact, frozen multi-scale feature space. Each emerging class is then independently modeled by a lightweight autoencoder, preserving previously learned representations without historical data replay or shared parameter updates.
During inference, \methodname{} directly recognizes all encountered classes by comparing their reconstruction errors, eliminating the need for a shared discriminative classifier. This design enables memory-efficient class expansion while avoiding interference with previously learned model parameters and the overhead of high-dimensional statistical matrices.

The main contributions of this work are summarized as follows:
\begin{itemize}

\item We propose \methodname{}, a replay-free TSCIL framework that decouples temporal-spectral feature extraction from class-specific modeling, enabling continuous acquisition of emerging time series classes without revisiting historical data or updating previously learned models.

\item We develop a compact, frozen temporal-spectral representation that captures complementary temporal dynamics and frequency patterns across multiple scales, providing a stable feature space for learning evolving time series classes.

\item We introduce a class-specific generative recognition strategy that models complex intra-class temporal variations through independent autoencoders, enabling incremental class expansion via reconstruction-based inference without shared classifier updates or large statistical matrices.

\item Extensive experiments across five time-series benchmarks demonstrate the effectiveness and memory efficiency of \methodname{}. Compared with the second-best method, our method improves average accuracy by $5.01\%$ while reducing retained memory by $98.26\%$. Our analyses further highlight the importance of temporal-spectral representations and nonlinear class-specific modeling for continual time-series recognition.
\end{itemize}

\section{Related Work}

\begin{figure*}[t]
  \centering
  \includegraphics[width=0.98\textwidth]{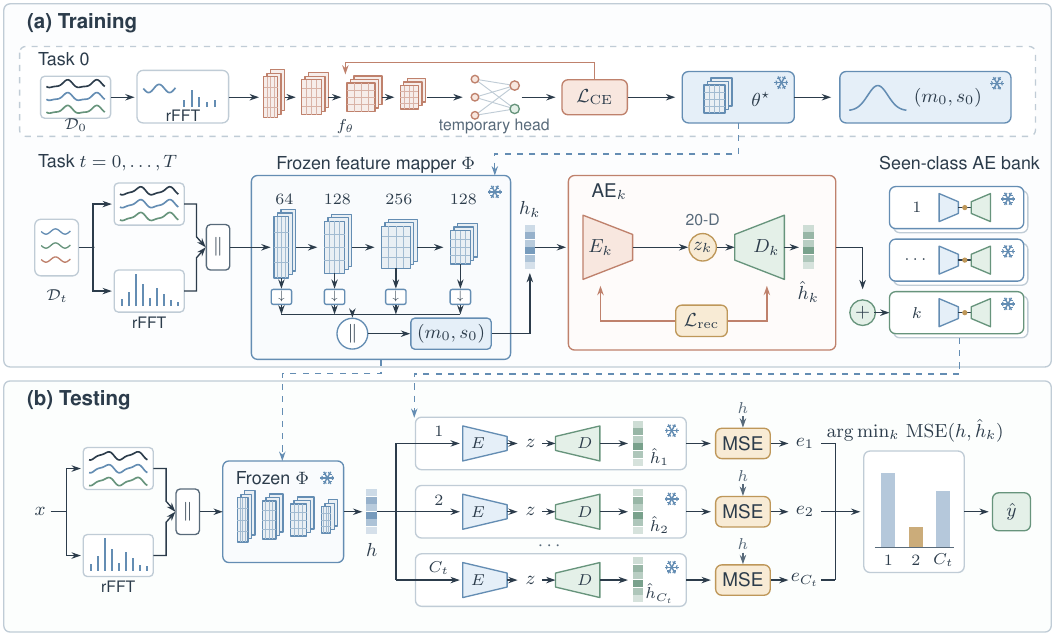}
  \caption{\methodname{} uses deterministic class-specific autoencoders. (a) Task 0 supervises the temporal-spectral CNN. The temporary head is discarded, and the CNN and Task-0 feature statistics are frozen. Each incoming class fits a bottleneck AE with reconstruction loss, then freezes it without replay. (b) Queries are encoded once and reconstructed by all seen-class AEs. The smallest feature MSE determines the prediction. Snowflakes mark frozen components.}
  \Description{Two-panel \methodnameplain{} AE architecture. Training combines raw time series and rFFT magnitudes, supervises a four-block CNN on Task 0, and freezes the multi-scale representation and Task-0 standardization. Each class has a deterministic encoder, a 20-dimensional bottleneck and a decoder trained by reconstruction loss. No variance branch, sampling or KL term is used. Trained class AEs are frozen and stored. Testing compares deterministic reconstructions from all seen-class AEs with the same query feature and predicts the minimum-MSE class. Waveforms and score bars are schematic.}
  \label{fig:architecture}
\end{figure*}

\subsection{Time Series Classification}

\begingroup
Time series classification (TSC) assigns semantic labels to sequential observations and supports applications such as activity recognition, physiological monitoring, and industrial sensing~\cite{ismail2019deep,foumani2024survey}. Existing research spans distance-based, dictionary-based, shapelet-based, and ensemble approaches, which characterize temporal patterns through complementary representations~\cite{schafer2017fast,middlehurst2021hive,middlehurst2024bakeoff}. Deep learning further enables representation learning through convolutional, recurrent, and attention-based architectures~\cite{ismail2020inceptiontime,wu2022timesnet,foumani2024survey}, while random convolutional transformations offer efficient alternatives~\cite{dempster2020rocket,dempster2021minirocket}. More recent studies explore self-supervised learning, time-frequency consistency, and large-scale pretraining to improve representation transfer across datasets and tasks~\cite{yue2022ts2vec,zhang2022tfc,goswami2024moment,liang2024foundation}. Despite these advances, conventional offline classification protocols generally assume a fixed label space, leaving continual acquisition of emerging classes and preservation of historical knowledge to be addressed.
\par\endgroup

\subsection{Time Series Class-Incremental Learning}

\begingroup
Time Series Class-Incremental Learning (TSCIL) learns new classes sequentially while retaining recognition of all seen classes without task identities~\cite{qiao2024class}. CIL research has largely focused on computer vision~\cite{delange2021continual,wang2024continual}. Regularization and distillation preserve important parameters or previous predictions~\cite{li2018lwf,aljundi2018mas}, whereas exemplar replay retains old samples for rehearsal~\cite{rebuffi2017icarl,buzzega2020dark}. To reduce historical-data storage and privacy concerns, generative replay replaces exemplars with synthesized samples~\cite{shin2017continual,ostapenko2019learning}. However, image-oriented generative replay methods struggle to capture complex temporal dependencies, leading to limited sample diversity and degraded recognition performance on multivariate time series~\cite{qiao2024class}.

Time-series methods have explored signal-specific memory management~\cite{kiyasseh2021clops,kwon2021fasticarl}, temporal alignment and confidence-aware distillation~\cite{qiao2023dt2w,liu2026cmdpcl}, and feature-space inversion~\cite{li2025anchorinv}. Analytical methods further use closed-form classifier updates to avoid storing exemplars~\cite{li2025tsacl,liu2026hada}. However, shared-model updates can interfere with old knowledge, while high-dimensional analytical heads incur dense-matrix storage costs. Subject-dependent within-class variations further complicate knowledge retention, since preserving only a few representative patterns may not adequately cover the diversity of each class~\cite{qiao2024class}. To address these trade-offs, \methodname{} performs replay-free generative classification using a frozen temporal-spectral representation and independently preserved class AEs, assigning labels by deterministic reconstruction error.
\par\endgroup

\section{Methodology}

We implement the concept of decoupled generative recognition through \methodname{}. The core philosophy of this framework is to avoid continually updating a shared discriminative head, which is vulnerable to catastrophic forgetting. Instead, we establish a universally frozen feature space and model each class with an independent autoencoder. As illustrated in Figure~\ref{fig:architecture}, \methodname{} decomposes replay-free time series continual recognition into three distinct stages: (1) constructing and freezing a compact temporal-spectral feature mapping $\Phi$; (2) independently fitting and freezing a lightweight autoencoder (AE) for each incoming class; and (3) evaluating all seen classes via deterministic feature reconstruction errors.

In the following, we define the continual learning problem setting in \S \ref{prob_sec}. In \S \ref{feat_sec}, we present the pipeline for extracting the frozen temporal-spectral representation, which serves as our universal semantic substrate. Given this stable foundation, we introduce the isolated per-class generative modeling and our novel deterministic scoring mechanism in \S \ref{gen_sec}. Finally, we provide the comprehensive workflow of \methodname{} in \S \ref{workflow_sec}.

\subsection{Problem Definition} \label{prob_sec}

Let a continual learner receive sequential task datasets $\mathcal{D}_0, \dots, \mathcal{D}_T$, where each task $t$ introduces a set of new classes $\mathcal{K}_t$ such that $\mathcal{K}_s \cap \mathcal{K}_t = \emptyset$ for $s \neq t$. Following task $t$, the cumulative set of seen classes is defined as:
\begin{equation}
\mathcal{C}_t = \bigcup_{s=0}^{t} \mathcal{K}_s, \qquad C_t = |\mathcal{C}_t|.
\end{equation}
Each task dataset $\mathcal{D}_t=\mathcal{F}_t\cup\mathcal{V}_t$ comprises disjoint fitting and early-stopping splits, respectively. During task $t$, the learner only has access to authentic data samples from $\mathcal{D}_t$. At inference, the objective is to predict labels across all previously seen classes $\mathcal{C}_t$ without relying on ground-truth task identities. Each input sequence is denoted as $x \in \mathbb{R}^{L \times M}$, characterized by a temporal length $L$ and sensor channels $M$.

\subsection{Constructing the Universal Semantic Substrate} \label{feat_sec}

To support class-specific generative modeling, the feature space must remain compact while preserving the diverse characteristics of multivariate time series. Temporal waveforms capture local dynamics, while frequency spectra provide complementary information about periodicity and global spectral structure. To effectively integrate these characteristics, we employ a multi-scale convolutional extractor that aggregates temporal and spectral features across different receptive fields, producing a compact and stable representation for incremental learning.

\paragraph{Temporal-spectral input construction}
For each sample $x$ and independently across each channel along the temporal axis, we compute the single-sided orthonormal real Fast Fourier Transform ($\operatorname{rFFT}$). We extract its magnitude spectrum and concatenate it with the raw signal $x$ prior to input normalization:
\begin{equation}
u(x) = \operatorname{Norm}_{\mathrm{in}}\!\left(x \;\Vert_{\mathrm{time}}\; |\operatorname{rFFT}(x)|\right) \in \mathbb{R}^{(L + \lfloor L/2 \rfloor + 1) \times M}.
\end{equation}
$\operatorname{Norm}_{\mathrm{in}}$ follows TSCIL's original input normalization configuration~\cite{qiao2024class}: LayerNorm across channels per position (HAR, DailySports, GrabMyo), InstanceNorm over concatenated time per channel (UWave), or identity (WISDM). Both use per-sample statistics without affine parameters or running statistics.

\paragraph{Multi-scale convolutional projection}
The augmented input $u(x)$ is processed by a 1D convolutional network comprising four blocks (Conv1d with kernel size $5$, ReLU, normalization, MaxPool2, and dropout). To comprehensively capture multi-scale temporal-spectral patterns, we apply global average pooling ($\operatorname{GAP}$) to the output of each block $\ell \in \{1, 2, 3, 4\}$ and concatenate the resulting vectors:
\begin{equation}
v_\theta(x) = \bigoplus_{\ell=1}^{4} \operatorname{GAP}\!\left(f_\theta^{(\ell)}(u(x))\right) \in \mathbb{R}^d,
\end{equation}
where our default channel widths $(64, 128, 256, 128)$ yield a compact multi-scale feature representation of dimension $d$.

\paragraph{Initial-task supervision}
We train the temporal-spectral CNN with a temporary linear classification head on the initial-task fitting split $\mathcal{F}_0$ using cross-entropy loss, selecting the checkpoint on $\mathcal{V}_0$. After training, we discard the classification head and permanently freeze the backbone parameters $\theta^\star$. The backbone remains unchanged throughout subsequent incremental tasks.

\paragraph{Frozen feature standardization}
We compute the coordinate-wise mean $m_0 \in \mathbb{R}^d$ and population standard deviation $s_0 \in \mathbb{R}^d$ using only fitting features from $\mathcal{F}_0$. These statistics define the frozen feature mapping $\Phi$:
\begin{equation}
h = \Phi(x) = \frac{v_{\theta^\star}(x) - m_0}{\max(s_0, \epsilon)}, \qquad \epsilon = 10^{-3}.
\end{equation}
Both the parameters $\theta^\star$ and the standardization statistics $(m_0, s_0)$ remain permanently fixed after $\mathcal{D}_0$. This absolute rigidity ensures zero representation drift across all subsequent tasks.

\subsection{Cross-Manifold Generative Inference} \label{gen_sec}

Given the frozen feature mapping $\Phi$, \methodname{} models each class independently to incorporate new knowledge without modifying previously learned models. Rather than updating a shared discriminative head, we train a lightweight autoencoder for each incoming class and perform classification by comparing class-specific reconstruction errors.

\paragraph{Independent class-specific AE training}
When a new class $k \in \mathcal{K}_t$ arrives, we extract its frozen training features $\mathcal{H}_k = \{\Phi(x) \mid (x, y) \in \mathcal{F}_t, y = k\}$. Instead of updating a shared discriminative head, we train an entirely independent, symmetric bottleneck AE ($d \to 128 \to r \to 128 \to d$, with bottleneck dimension $r = 20$) consisting of a deterministic encoder $E_k$ and decoder $D_k$ tailored exclusively to this class. Its reconstruction mapping is $A_k=D_k\circ E_k$. The model parameters are optimized by minimizing the reconstruction objective over $\mathcal{H}_k$:
\begin{equation}
\mathcal{L}_k = \mathbb{E}_{h \in \mathcal{H}_k}\!\left[\frac{1}{2}\left\|h - D_k(E_k(h))\right\|_2^2\right]\!.
\label{eq:ae_loss}
\end{equation}
We restore the checkpoint of $A_k$ minimizing reconstruction loss on $\mathcal{H}_k^{\mathrm{val}}=\{\Phi(x)\mid(x,y)\in\mathcal{V}_t, y=k\}$, then permanently freeze $E_k$ and $D_k$. Crucially, historical class AEs are never updated, accessed, or sampled for replay during this process.

\paragraph{Deterministic scoring and invariance}
At inference, each seen-class AE deterministically reconstructs the extracted feature $\Phi(x)$ in a single forward pass without Monte Carlo sampling. We evaluate each seen class $k \in \mathcal{C}_t$ by its dimension-averaged reconstruction error and predict the class with the smallest score:
\begin{equation}
e_k(x) = \frac{1}{d}\left\|\Phi(x) - D_k\!\left(E_k(\Phi(x))\right)\right\|_2^2, \qquad \hat{y} = \underset{k \in \mathcal{C}_t}{\arg\min}\; e_k(x).
\label{eq:recon_score}
\end{equation}
Because the mapping $\Phi$, the statistics $(m_0, s_0)$, and all historical AEs remain strictly frozen, the scoring function of any previously learned class $k \in \mathcal{C}_t$ is invariant across future tasks: $e_k^{(t+1)}(x) = e_k^{(t)}(x)$. This score-function invariance prevents parameter drift in historical class models without updating past parameters, while newly added class models still participate in candidate score competition at inference.

\subsection{Workflow of \methodnameplain{}}
\label{workflow_sec}

We summarize the overall training and inference workflow of \methodname{} in Algorithm~\ref{alg:overview}, which separates frozen feature extraction from independent class-specific generative modeling. The frozen AE bank after task $t$ is $\mathcal{B}_t=\{(k,A_k)\mid k\in\mathcal{C}_t\}$.

\paragraph{Architectural design}
\methodname{} consists of a shared feature extractor and a collection of independent class-specific AEs. The feature extractor employs a lightweight four-block 1D CNN to capture multi-scale patterns from temporal signals augmented with $\operatorname{rFFT}$ spectral information. Each class is modeled by a compact AE with a low-dimensional bottleneck, enabling class-specific feature reconstruction without updating a shared discriminative head.

\paragraph{Base training ($\mathcal{D}_0$)}
In \methodname{}, the feature extractor and a temporary classification head are trained on $\mathcal{F}_0$ using cross-entropy loss, with checkpoint selection on $\mathcal{V}_0$. After training, the classification head is discarded, and the backbone parameters $\theta^\star$ are frozen. The feature standardization statistics $(m_0,s_0)$ are then computed from the fitting features of $\mathcal{F}_0$ and fixed for all subsequent tasks. Finally, an independent AE is trained and frozen for each class in $\mathcal{D}_0$ using the resulting feature representations.

\paragraph{Incremental learning ($\mathcal{D}_t$, $t\geq 1$)}
For each subsequent task, incoming samples are mapped into the fixed feature space through $\Phi$. An independent AE is initialized and trained for each newly introduced class using Eq.~\eqref{eq:ae_loss}. Once trained, its parameters are frozen and the model is added to the collection of previously learned class models.\space\space
With the default $576$-dimensional features, each new class incurs a modest storage overhead of approximately $0.5853$~MiB~(FP32) for its $153{,}428$ AE parameters. Throughout this process, neither the feature extractor nor any existing AE is updated, and no historical samples or generated replay data are required.

\paragraph{Inference}
Given a test sequence $x$, the frozen feature extractor produces its representation $h=\Phi(x)$. This representation is evaluated by all AEs corresponding to previously encountered classes. Each AE deterministically reconstructs $h$ via $D_k(E_k(h))$, and the predicted label is assigned to the class with the smallest reconstruction error according to Eq.~\eqref{eq:recon_score}. This procedure enables classification across all learned classes without task-identity information, historical data replay, or updates to existing models.

\begin{algorithm}[H]
\caption{\methodnameplain{}: Training and inference}
\label{alg:overview}
\normalsize\raggedright
\textbf{Require:} Task-specific fitting and early-stopping
splits $\mathcal{F}_t,\mathcal{V}_t$ and new class sets
$\mathcal{K}_t$, for $t=0,\ldots,T$.\par
\begin{pseudocode}
\pstep{Train the CNN feature extractor and a temporary
classification head on $\mathcal{F}_0$; select the best
checkpoint using $\mathcal{V}_0$.}
\pstep{Discard the head; freeze $\theta^\star$;
compute $(m_0,s_0)$ from $\mathcal{F}_0$.}
\pstep{Fix $\Phi(x)=(v_{\theta^\star}(x)-m_0)/
\max(s_0,\epsilon)$, where $\epsilon=10^{-3}$
and operations are element-wise.}
\pstep{Initialize $\mathcal{B}_{-1}\gets\varnothing$.}
\pstep{\textbf{for} task $t=0,\ldots,T$ \textbf{do}}
\pstep[1]{$\mathcal{B}_t\gets\mathcal{B}_{t-1}$.}
\pstep[1]{\textbf{for each} new class
$k\in\mathcal{K}_t$ \textbf{do}}
\pstep[2]{Construct
$\mathcal{H}_k=\{\Phi(x):(x,y)\in\mathcal{F}_t,\ y=k\}$.}
\pstep[2]{Construct
$\mathcal{H}_k^{\mathrm{val}}
=\{\Phi(x):(x,y)\in\mathcal{V}_t,\ y=k\}$.}
\pstep[2]{Train a new AE $A_k=D_k\circ E_k$
on $\mathcal{H}_k$ using Eq.~\eqref{eq:ae_loss}.}
\pstep[2]{Restore the best checkpoint on
$\mathcal{H}_k^{\mathrm{val}}$; freeze $A_k$.}
\pstep[2]{$\mathcal{B}_t\gets
\mathcal{B}_t\cup\{(k,A_k)\}$.}
\pstep[1]{\textbf{end for}}
\pstep[1]{Discard current-task data, extracted
features, and training optimizers.}
\pstep{\textbf{end for}}
\pstep{\textbf{Inference:} Given a query $x$ after
task $t\in\{0,\ldots,T\}$, retrieve
$\mathcal{B}_t$ and compute $h\gets\Phi(x)$.}
\pstep{\textbf{return}
$\hat y=\arg\min_{k\in\mathcal{C}_t}
\frac{1}{d}\|h-A_k(h)\|_2^2$.}
\end{pseudocode}
\end{algorithm}

\section{Experiments}

\begin{table}[t]
  \caption{Benchmark class counts, task horizons (2 classes per task), and input dimensions ($L \times M$) across the five TSCIL benchmarks compared with the original benchmark~\cite{qiao2024class}.}
  \label{tab:datasets}
  \centering
  \footnotesize
  \begin{tabular}{lrrrr}
    \toprule
    \textbf{Dataset} & \textbf{Shape ($L \times M$)} & \textbf{Classes} & \textbf{Orig.\ Tasks} & \textbf{Our Tasks} \\
    \midrule
    HAR~\cite{anguita2013har} & $128 \times 9$ & 6 & 3 & 3 \\
    UWave~\cite{liu2009uwave} & $315 \times 3$ & 8 & 4 & 4 \\
    DailySports~\cite{altun2010har} & $125 \times 45$ & 18 & 6 & 9 \\
    WISDM~\cite{weiss2019wisdm} & $200 \times 3$ & 18 & 6 & 9 \\
    GrabMyo~\cite{pradhan2022grabmyo} & $128 \times 28$ & 16 & 5 & 8 \\
    \bottomrule
  \end{tabular}
\end{table}

\begin{table*}[t]
  \caption{Final test accuracy (mean $\pm$ SD, \%) and estimated retained memory (MiB) over five class orders and five seeds per dataset (25 runs). Macro averages weight datasets equally. Offline denotes all-class joint training. \textbf{Bold} and \underline{underlining} indicate the highest and second-highest five-dataset macro-average accuracy among continual-learning methods, respectively.}
  \label{tab:main_5x5}
  \centering
  \footnotesize
  \setlength{\tabcolsep}{3.0pt}
  \begin{tabular}{l cc cc cc cc cc cc}
    \toprule
    & \multicolumn{2}{c}{\textbf{HAR}} & \multicolumn{2}{c}{\textbf{UWave}} & \multicolumn{2}{c}{\textbf{DailySports}} & \multicolumn{2}{c}{\textbf{WISDM}} & \multicolumn{2}{c}{\textbf{GrabMyo}} & \multicolumn{2}{c}{\textbf{5-Dataset Macro}} \\
    \cmidrule(lr){2-3} \cmidrule(lr){4-5} \cmidrule(lr){6-7} \cmidrule(lr){8-9} \cmidrule(lr){10-11} \cmidrule(lr){12-13}
    \textbf{Method} & Acc (\%) & MiB & Acc (\%) & MiB & Acc (\%) & MiB & Acc (\%) & MiB & Acc (\%) & MiB & \cellcolor{accbluebg}Acc (\%) & \cellcolor{mibgray}MiB \\
    \midrule
    \multicolumn{13}{l}{\emph{Reference Bounds (Joint-Training Reference \& Fine-Tuning Lower Bound)}} \\
    Offline (Joint Ref.) & $92.06 \pm 1.30$ & 1.43 & $95.02 \pm 0.93$ & 1.42 & $98.50 \pm 0.52$ & 1.48 & $85.83 \pm 2.37$ & 1.43 & $87.73 \pm 1.61$ & 1.45 & \cellcolor{accbluebg}\textcolor{accbluetext}{91.83} & \cellcolor{mibgray}1.44 \\
    Naive & $38.23 \pm 8.72$ & 1.43 & $24.62 \pm 0.29$ & 1.42 & $15.14 \pm 3.43$ & 1.48 & $11.40 \pm 1.20$ & 1.43 & $12.16 \pm 0.14$ & 1.45 & \cellcolor{accbluebg}\textcolor{accbluetext}{20.31} & \cellcolor{mibgray}1.44 \\
    \midrule
    \multicolumn{13}{l}{\emph{Experience \& Generative Replay Methods}} \\
    ER~\cite{rolnick2019experience} & $89.33 \pm 2.05$ & 3.04 & $79.11 \pm 3.37$ & 1.56 & $96.19 \pm 1.68$ & 8.43 & $46.14 \pm 6.12$ & 3.49 & $51.44 \pm 1.23$ & 23.78 & \cellcolor{accbluebg}\textcolor{accbluetext}{72.44} & \cellcolor{mibgray}8.06 \\
    DER~\cite{buzzega2020dark} & $89.95 \pm 1.17$ & 3.05 & $75.98 \pm 4.94$ & 1.57 & $97.07 \pm 0.95$ & 8.45 & $32.48 \pm 6.25$ & 3.55 & $53.81 \pm 1.37$ & 23.88 & \cellcolor{accbluebg}\textcolor{accbluetext}{69.86} & \cellcolor{mibgray}8.10 \\
    iCaRL~\cite{rebuffi2017icarl} & $88.51 \pm 2.13$ & 4.65 & $84.85 \pm 1.90$ & 1.71 & $92.71 \pm 1.56$ & 15.39 & $33.20 \pm 3.30$ & 5.55 & $51.15 \pm 1.02$ & 46.09 & \cellcolor{accbluebg}\textcolor{accbluetext}{70.09} & \cellcolor{mibgray}14.68 \\
    ASER~\cite{shim2021online} & $88.59 \pm 2.10$ & 3.04 & $76.57 \pm 2.68$ & 1.56 & $96.06 \pm 1.24$ & 8.43 & $33.17 \pm 2.76$ & 3.49 & $48.76 \pm 1.19$ & 23.78 & \cellcolor{accbluebg}\textcolor{accbluetext}{68.63} & \cellcolor{mibgray}8.06 \\
    CLOPS~\cite{kiyasseh2021clops} & $86.28 \pm 3.07$ & 3.04 & $59.60 \pm 11.88$ & 1.56 & $77.51 \pm 6.64$ & 8.43 & $22.01 \pm 1.71$ & 3.49 & $35.52 \pm 1.31$ & 23.78 & \cellcolor{accbluebg}\textcolor{accbluetext}{56.19} & \cellcolor{mibgray}8.06 \\
    FastICARL~\cite{kwon2021fasticarl} & $86.58 \pm 1.79$ & 3.04 & $82.72 \pm 2.31$ & 1.57 & $88.78 \pm 3.14$ & 8.44 & $30.28 \pm 2.79$ & 3.50 & $47.46 \pm 1.25$ & 23.79 & \cellcolor{accbluebg}\textcolor{accbluetext}{67.16} & \cellcolor{mibgray}8.07 \\
    GR~\cite{shin2017continual,qiao2024class} & $51.44 \pm 8.31$ & 46.55 & $73.90 \pm 7.99$ & 68.61 & $31.79 \pm 4.46$ & 393.72 & $12.82 \pm 1.53$ & 46.72 & $12.17 \pm 0.15$ & 178.50 & \cellcolor{accbluebg}\textcolor{accbluetext}{36.42} & \cellcolor{mibgray}146.82 \\
    \midrule
    \multicolumn{13}{l}{\emph{Regularization, Distillation, and Replay-Free Adaptation Methods}} \\
    LwF~\cite{li2018lwf} & $44.74 \pm 7.47$ & 1.43 & $24.87 \pm 0.54$ & 1.42 & $15.54 \pm 3.48$ & 1.48 & $12.10 \pm 1.79$ & 1.43 & $12.20 \pm 0.19$ & 1.45 & \cellcolor{accbluebg}\textcolor{accbluetext}{21.89} & \cellcolor{mibgray}1.44 \\
    MAS~\cite{aljundi2018mas} & $51.00 \pm 6.19$ & 2.85 & $25.07 \pm 1.60$ & 2.84 & $24.55 \pm 2.75$ & 2.95 & $11.21 \pm 1.35$ & 2.85 & $10.83 \pm 0.50$ & 2.91 & \cellcolor{accbluebg}\textcolor{accbluetext}{24.53} & \cellcolor{mibgray}2.88 \\
    DT2W~\cite{qiao2023dt2w} & $58.43 \pm 8.36$ & 1.43 & $50.11 \pm 8.88$ & 1.42 & $14.19 \pm 2.64$ & 1.49 & $11.14 \pm 0.80$ & 1.43 & $13.06 \pm 0.49$ & 1.46 & \cellcolor{accbluebg}\textcolor{accbluetext}{29.38} & \cellcolor{mibgray}1.45 \\
    LDC~\cite{gomezvilla2024ldc} & $66.95 \pm 7.55$ & 1.45 & $67.10 \pm 2.27$ & 1.45 & $28.12 \pm 4.62$ & 1.55 & $21.99 \pm 1.44$ & 1.50 & $35.83 \pm 0.97$ & 1.52 & \cellcolor{accbluebg}\textcolor{accbluetext}{44.00} & \cellcolor{mibgray}1.49 \\
    PGM~\cite{wan2025pgm} & $39.80 \pm 9.64$ & 24.12 & $46.11 \pm 6.63$ & 31.40 & $13.96 \pm 3.79$ & 70.92 & $11.49 \pm 1.31$ & 68.43 & $20.19 \pm 1.59$ & 62.35 & \cellcolor{accbluebg}\textcolor{accbluetext}{26.31} & \cellcolor{mibgray}51.44 \\
    AnchorInv~\cite{li2025anchorinv} & $44.49 \pm 8.73$ & 2.24 & $26.79 \pm 2.39$ & 2.38 & $27.69 \pm 5.26$ & 10.07 & $13.10 \pm 1.99$ & 3.12 & $14.24 \pm 0.84$ & 6.60 & \cellcolor{accbluebg}\textcolor{accbluetext}{25.26} & \cellcolor{mibgray}4.88 \\
    \midrule
    \multicolumn{13}{l}{\emph{Analytical Continual Learning Methods (Replay-Free)}} \\
    KLDA~\cite{momeni2025klda} & $87.39 \pm 1.45$ & 405.36 & $94.53 \pm 0.33$ & 405.51 & $97.29 \pm 0.83$ & 406.32 & $79.68 \pm 0.98$ & 406.27 & $62.14 \pm 1.30$ & 406.15 & \cellcolor{accbluebg}\textcolor{accbluetext}{84.20} & \cellcolor{mibgray}405.92 \\
    TS-ACL~\cite{li2025tsacl} & $89.81 \pm 0.95$ & 525.23 & $93.83 \pm 0.21$ & 525.34 & $97.41 \pm 0.16$ & 526.00 & $80.22 \pm 1.27$ & 525.95 & $61.80 \pm 1.15$ & 525.86 & \cellcolor{accbluebg}\textcolor{accbluetext}{\underline{84.61}} & \cellcolor{mibgray}\underline{525.68} \\
    HADA~\cite{liu2026hada} & $89.82 \pm 2.42$ & 525.41 & $91.59 \pm 0.74$ & 525.53 & $98.24 \pm 0.81$ & 526.19 & $70.99 \pm 2.35$ & 526.14 & $63.07 \pm 1.37$ & 526.04 & \cellcolor{accbluebg}\textcolor{accbluetext}{82.74} & \cellcolor{mibgray}525.86 \\
    \midrule
    \textbf{Ours} & $90.06 \pm 1.10$ & 4.94 & $93.22 \pm 0.58$ & 6.10 & $98.52 \pm 0.26$ & 12.01 & $80.02 \pm 2.34$ & 11.96 & $82.46 \pm 0.90$ & 10.82 & \cellcolor{accbluebg}$\textcolor{accbluetext}{\textbf{88.85}}$ & \cellcolor{mibgray}\textbf{9.17} \\
    \bottomrule
  \end{tabular}
\end{table*}

\subsection{Experimental Setup}

\paragraph{Datasets.}
We evaluate \methodname{} on five multivariate time series benchmarks: HAR~\cite{anguita2013har}, UWave~\cite{liu2009uwave}, DailySports~\cite{altun2010har}, WISDM~\cite{weiss2019wisdm}, and GrabMyo~\cite{pradhan2022grabmyo}. Following the established TSCIL benchmark~\cite{qiao2024class}, each incremental task introduces two new classes. We include all available classes in the incremental stream, resulting in 3, 4, 9, 9, and 8 tasks, respectively. Dataset statistics and task configurations are summarized in Table~\ref{tab:datasets}.

\paragraph{Evaluation Metrics.}
We evaluate classification accuracy after each incremental task and report final test accuracy after all classes have been introduced. To assess memory efficiency, we measure the retained state (MiB) required for continued incremental learning, including model parameters, stored exemplars, normalization statistics, analytical matrices, and persistent optimizer states.\space\space
All memory components are measured using their native data types. We further examine the accuracy--memory trade-off to evaluate recognition performance under storage constraints. 

\paragraph{Implementation Details.}
We implement all methods within a unified evaluation framework based on~\cite{qiao2024class}, using the standard four-block 1D CNN backbone while preserving method-specific representations and learning mechanisms. To prevent data leakage, the development data of each class is partitioned into training (80\%), early-stopping (10\%), and validation (10\%) subsets. Hyperparameters are determined using development data and fixed across all class orders before formal evaluation. The validation data is then merged into the training set for final model fitting, while the early-stopping subset is retained for epoch selection. The official test sets are used exclusively for evaluation. Each configuration is evaluated over five class orders and five random seeds, yielding 25 runs per dataset. We average results across seeds within each order and compute 95\% confidence intervals across the five order-level means using Student's $t$-distribution ($\mathrm{df}=4$).

\paragraph{Baselines.}
We compare \methodname{} against representative TSCIL and conventional class-incremental learning methods under a unified evaluation protocol. The baselines include experience replay methods (ER~\cite{rolnick2019experience}, DER~\cite{buzzega2020dark}, iCaRL~\cite{rebuffi2017icarl}, ASER~\cite{shim2021online}, CLOPS~\cite{kiyasseh2021clops}, and FastICARL~\cite{kwon2021fasticarl}), generative replay (GR~\cite{shin2017continual,qiao2024class}), regularization-, distillation-, and adaptation-based methods (LwF~\cite{li2018lwf}, MAS~\cite{aljundi2018mas}, DT2W~\cite{qiao2023dt2w}, LDC~\cite{gomezvilla2024ldc}, PGM~\cite{wan2025pgm}, and AnchorInv~\cite{li2025anchorinv}), and analytical learning methods (KLDA~\cite{momeni2025klda}, TS-ACL~\cite{li2025tsacl}, and HADA~\cite{liu2026hada}). We additionally include \emph{Offline}, which jointly trains on all classes, as a reference, and \emph{Naive}, which sequentially fine-tunes on incoming tasks without any forgetting mitigation mechanism. For fair comparison, all methods follow a common four-block CNN backbone design while retaining their method-specific feature representations, classification heads, and incremental update strategies. Consistent dataset partitions, class orders, and evaluation metrics are applied across all methods.

\subsection{Main Performance}

\begin{figure}[!b]
  \centering
  \includegraphics[width=\columnwidth]{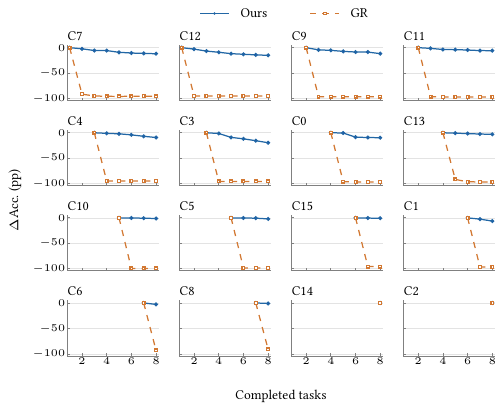}
  \caption{GrabMyo per-class test accuracy changes (pp) are measured from first evaluation and averaged over five seeds under a fixed class arrival sequence. Blue solid curves show Ours (AE), and orange dashed curves show GR.}
  \Description{Sixteen panels plot class-wise changes in test accuracy over eight tasks on GrabMyo under a fixed class arrival sequence. Ours uses deterministic class-specific AEs on frozen temporal-spectral features and retains more old-class accuracy than GR under this sequence, while still showing some decline as new classes arrive. The two classes introduced at the final task have only their initial zero-change point.}
  \label{fig:grabmyo_retention}
\end{figure}

Table~\ref{tab:main_5x5} and Figure~\ref{fig:fig1_tradeoff} summarize the overall performance and accuracy--memory trade-off across five TSCIL benchmarks. \methodname{} achieves the highest five-dataset average final accuracy of $88.85\%$ among all evaluated continual-learning methods, with only $9.17$~MiB of retained memory. It achieves the best accuracy on HAR, DailySports, and GrabMyo, while remaining competitive with KLDA on UWave and TS-ACL on WISDM.
Exemplar-based methods, such as ER and DER, achieve reasonable performance on HAR and DailySports but exhibit substantial accuracy degradation on WISDM and GrabMyo. This suggests that limited historical exemplars may be insufficient to capture complex intra-class temporal variations. Regularization- and distillation-based methods, such as LwF and DT2W, generally achieve lower accuracy, reflecting their susceptibility to catastrophic forgetting. Generative replay (GR) avoids storing historical samples but performs poorly across several datasets, potentially due to limited synthesis fidelity and accumulated errors across incremental tasks. Its additional generator and optimizer states also introduce considerable memory overhead.
Analytical methods, such as KLDA and TS-ACL, demonstrate strong classification performance, achieving the highest accuracy among continual-learning methods on UWave and WISDM, respectively. However, high-dimensional feature expansion and the associated statistical matrices can introduce substantial retained memory overhead, exceeding $400$~MiB for some evaluated configurations. Their comparatively lower performance on GrabMyo suggests that increasing feature dimensionality alone may be insufficient to characterize complex nonlinear intra-class variations.

In contrast, \methodname{} combines complementary temporal-spectral information with independent nonlinear class modeling, enabling effective recognition of diverse temporal patterns without maintaining large analytical matrices. Compared with TS-ACL, the strongest analytical baseline in five-dataset average accuracy, \methodname{} achieves a $5.01\%$ relative improvement in average accuracy while reducing retained memory by approximately $98.26\%$ ($9.17$ versus $525.68$~MiB). This favorable trade-off highlights the effectiveness of preserving class-specific knowledge within a compact, frozen feature space rather than relying on high-dimensional statistical expansion.

To further examine knowledge preservation during incremental learning, Figure~\ref{fig:grabmyo_retention} tracks per-class accuracy changes from their first evaluation on GrabMyo. While generative replay (GR) suffers severe degradation as new classes arrive, \methodname{} maintains substantially more stable historical recognition, with a mean old-class accuracy decline of only $7.17$~pp, compared with $95.73$~pp for GR. This advantage is consistent with our independent class modeling strategy, which preserves previously learned scoring functions within a fixed temporal-spectral feature space without updating historical parameters or relying on replay.

\begin{figure*}[t]
  \centering
  \includegraphics[width=\textwidth]{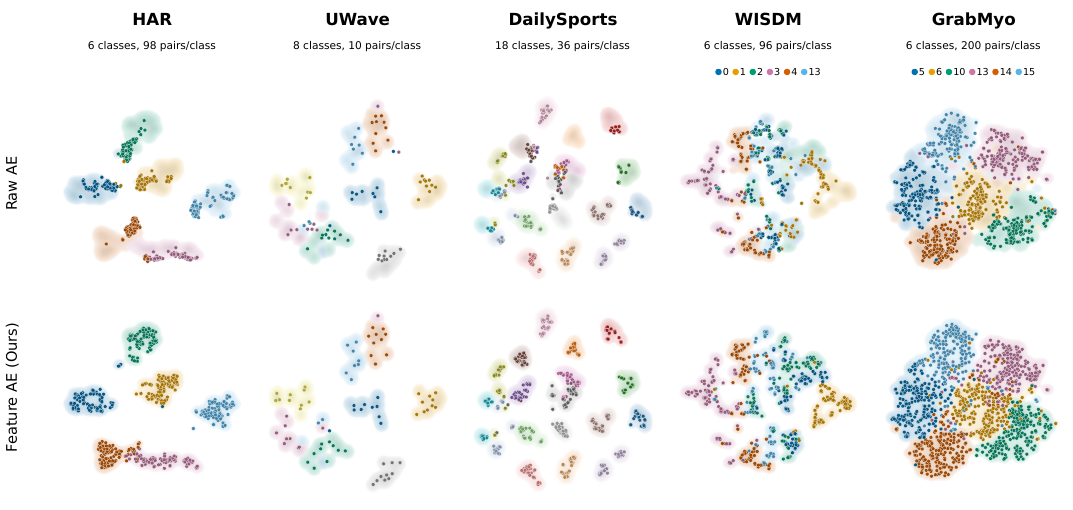}
  \caption{Raw-series AE (top) and feature AE (ours, bottom) deterministically reconstruct identical validation inputs. Shading shows real density, and dots show paired reconstructions in a shared joint t-SNE diagnostic space per column. All classes are shown for HAR/UWave/DailySports. Six classes selected post hoc are shown for WISDM/GrabMyo.}
  \Description{A two-row, five-column figure. Columns are HAR, UWave, DailySports, WISDM and GrabMyo. Deterministic raw-series AE reconstructions are above feature AE reconstructions. The first three columns show all 6, 8 and 18 classes, with 98, 10 and 36 paired validation inputs per class respectively. WISDM shows classes 0, 1, 2, 3, 4 and 13 with 96 inputs per class. GrabMyo shows classes 5, 6, 10, 13, 14 and 15 with 200 inputs per class. In every column the same real inputs, class colors, background density and joint t-SNE coordinates are shared by both rows. Each real input has exactly one deterministic reconstruction in each row. Raw reconstructions pass through the same frozen initial-task CNN and frozen all-class diagnostic MLP as the real inputs. Feature reconstructions pass through the same diagnostic MLP, which is used only for visualization. WISDM and GrabMyo classes were selected post hoc for neighborhood separation. All three other datasets use all classes.}
  \label{fig:vae_feature_mapping}
\end{figure*}

\subsection{Ablation Study}
\label{sec:ablation}

We examine the key design choices of \methodname{}, including feature-space reconstruction, temporal-spectral fusion, multi-scale feature extraction, and class-specific modeling. The results are summarized in Table~\ref{tab:mapping_ablation}.

\paragraph{Feature-Space Reconstruction.}
We first investigate the benefit of reconstructing frozen features rather than raw time series. While direct raw-series reconstruction (\emph{No Map}) performs well on HAR and UWave, its effectiveness deteriorates substantially on more complex datasets. In contrast, feature-space reconstruction (\emph{Base}) yields relative accuracy gains of $166.56\%$ on WISDM and $96.02\%$ on GrabMyo. By integrating temporal dynamics and spectral characteristics into a compact feature space, the frozen CNN provides more informative representations for class-specific reconstruction, particularly for complex multichannel signals. Figure~\ref{fig:vae_feature_mapping} further illustrates this difference, where feature-space reconstructions exhibit closer alignment with real class neighborhoods in the selected t-SNE visualizations. These findings support performing class-specific generative modeling in a learned temporal-spectral feature space rather than directly in the raw signal domain.

\paragraph{Temporal-Spectral Inputs.}
We investigate the contribution of spectral information by comparing time-domain inputs (\emph{Time}), channel-wise spectral concatenation (\emph{FFT-Ch}), and length-wise temporal-spectral fusion (\emph{rFFT-Len}). As shown in Table~\ref{tab:mapping_ablation}, spectral augmentation provides dataset-dependent improvements, with \emph{rFFT-Len} achieving a $24.44\%$ relative accuracy gain over \emph{Time} on GrabMyo. This improvement highlights the complementary role of spectral information, which exposes global frequency characteristics alongside local temporal variations, benefiting complex biosignal recognition. Although other configurations perform better on individual datasets, \emph{rFFT-Len} achieves the highest average accuracy across all five benchmarks while preserving the original sensor channel structure. These results motivate its adoption as the default temporal-spectral input design in \methodname{}.

\paragraph{Multi-Scale Features.}
We further examine the effect of representation structure and capacity by comparing last-layer features (\emph{Last}, $128$-D), multi-scale concatenation (\emph{Base}, $576$-D), and a wider backbone (\emph{Wide}, $1152$-D). Multi-scale features consistently outperform the last-layer representation, with a $34.17\%$ relative accuracy gain on GrabMyo. By aggregating intermediate convolutional features, the multi-scale representation preserves information across different receptive fields, allowing both localized signal variations and broader temporal-spectral structures to contribute to recognition. Further increasing the feature dimension provides only modest and dataset-dependent improvements at additional model cost. We therefore adopt the compact $576$-D multi-scale representation to balance recognition performance and memory efficiency.

\paragraph{Class-Specific Models.}
To assess the importance of nonlinear class-specific modeling, we compare deterministic AEs with class centroids (\emph{Mean}), diagonal Gaussian estimators (\emph{Diag.\ Gauss.}), rank-$20$ PCA, and variational autoencoders (\emph{VAE}) under identical frozen representations. Simple statistical models consistently underperform AEs, while PCA remains competitive on simpler datasets but struggles with complex temporal signals. In particular, AEs outperform PCA by a relative $48.07\%$ on WISDM and $44.34\%$ on GrabMyo. These results suggest that even within a shared temporal-spectral feature space, simple statistical models and linear subspaces may not adequately capture nonlinear intra-class variations. In contrast, class-specific AEs provide more flexible reconstruction models for such complex feature distributions. Deterministic AEs also achieve competitive performance relative to VAEs across the benchmarks, without requiring stochastic latent sampling or variational regularization. These findings motivate lightweight deterministic AEs for independent class-specific recognition.

\subsection{Generalization of Frozen Features}
\label{sec:pretraining_coverage}

Beyond architectural design, we investigate how pretraining coverage affects the generalization of frozen temporal-spectral representations to emerging classes.

\paragraph{Pretraining Coverage.}
We vary the number $N$ of tasks used for CNN pretraining before freezing the backbone. For $N=0$, the CNN is randomly initialized and immediately frozen without gradient updates. The $N=1$ configuration is standard \methodname{}, with supervision restricted to the initial task. For expanded coverage ($N=2,4$), the CNN is pretrained on the first $N$ tasks ($2N$ classes). We also include all-class pretraining. In every configuration, feature standardization uses only the initial-task training features, and the sequential class-specific AE fitting protocol is retained.\space\space
Expanded and all-class pretraining use future-task labels and are therefore evaluated as representation diagnostics.

Table~\ref{tab:pretrain_coverage} summarizes the final test accuracy under these configurations. Overall, broader pretraining coverage generally improves recognition performance, highlighting the importance of learning transferable representations. This trend is particularly evident on GrabMyo, where increasing coverage from $N=0$ to $N=4$ yields a $23.74\%$ relative accuracy gain, with further gains under all-class supervision. These results suggest that exposure to more diverse classes helps the temporal-spectral backbone learn representations that better support subsequent class-specific reconstruction.\space\space
Interestingly, under all-class pretraining, All AE achieves higher mean accuracy than the corresponding linear classification head on all datasets. This consistent advantage further supports the effectiveness of our class-specific reconstruction strategy.

\begin{figure}[t]
  \centering
  \includegraphics[width=\columnwidth]{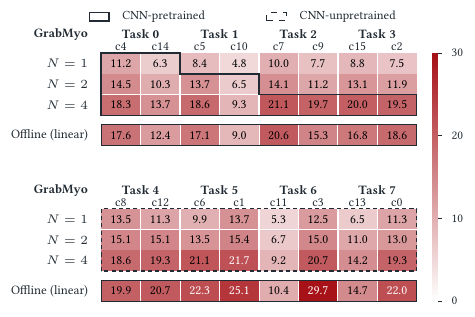}
  \caption{GrabMyo per-class accuracy gains over $N=0$ (pp) average five seeds under one class arrival sequence (Task 0: c4, c14). Solid/dashed outlines indicate CNN-pretrained/unpretrained classes. Offline is the all-class linear head. Darker red indicates larger gains.}
  \Description{Heatmap of final per-class AE accuracy gains relative to the N0 random frozen CNN plus AE baseline on GrabMyo under the class arrival sequence starting with Task 0 classes c4 and c14. Each cell averages five paired seeds. Rows N1, N2 and N4 retain deterministic per-class AEs. The separate Offline linear row uses the all-class CNN classifier head. The two panels preserve class arrival order. Solid outlines mark pretrained classes and the dashed region marks the fixed eight classes unpretrained under N up to 4. All 64 displayed mean gains are positive. A white-to-red sequential scale spans zero to 30 percentage points, with darker red indicating larger gains.}
  \label{fig:grabmyo_pretrain_heatmap}
  \vspace{-0.5cm}
\end{figure}

\paragraph{Transfer to Unseen Classes.}
We further investigate whether broader pretraining coverage benefits classes excluded from CNN pretraining. To isolate this effect, we evaluate fixed cohorts of unseen classes under different pretraining horizons ($N\leq4$). Broader pretraining coverage can also benefit unpretrained classes on complex datasets. For example, increasing coverage from $N=1$ to $N=4$ yields an $8.11$-percentage-point accuracy gain on GrabMyo. Figure~\ref{fig:grabmyo_pretrain_heatmap} further illustrates that improvements extend to classes not encountered during CNN pretraining. These findings suggest that broader supervision can enhance the transferability of frozen temporal-spectral features, allowing newly introduced classes to benefit from a more informative shared representation without updating the backbone. 

\begin{table}[t]
  \caption{Feature representation, input design, and class-model ablations report final test accuracy (mean $\pm$ SD, \%, $25$ runs per cell). Bold indicates the column best.}
  \label{tab:mapping_ablation}
  \centering
  \footnotesize
  \setlength{\tabcolsep}{1.5pt}
  \begin{tabular}{@{}l*{5}{r}@{}}
    \toprule
    \textbf{Variant} & \textbf{HAR} & \textbf{UWave} & \textbf{DailySp.} & \textbf{WISDM} & \textbf{GrabMyo} \\
    \midrule
    \multicolumn{6}{l}{\textit{Feature representation (AE)}} \\
    No Map & $\mathbf{94.43 \pm 0.48}$ & $\mathbf{96.00 \pm 0.25}$ & $97.35 \pm 0.71$ & $30.02 \pm 0.95$ & $42.07 \pm 0.70$ \\
    Last (128) & $84.01 \pm 2.41$ & $92.59 \pm 1.16$ & $96.17 \pm 1.34$ & $74.64 \pm 4.07$ & $61.46 \pm 3.65$ \\
    Base (576) & $90.06 \pm 1.10$ & $93.22 \pm 0.58$ & $98.52 \pm 0.26$ & $\mathbf{80.02 \pm 2.34}$ & $82.46 \pm 0.90$ \\
    Wide (1152) & $90.92 \pm 0.64$ & $93.32 \pm 0.65$ & $98.61 \pm 0.14$ & $79.20 \pm 2.34$ & $\mathbf{85.19 \pm 0.72}$ \\
    \midrule
    \multicolumn{6}{l}{\textit{Input design (AE)}} \\
    Time & $89.59 \pm 1.03$ & $92.74 \pm 0.76$ & $\mathbf{99.22 \pm 0.22}$ & $79.37 \pm 1.83$ & $66.27 \pm 1.08$ \\
    FFT-Ch & $90.83 \pm 1.12$ & $94.41 \pm 0.69$ & $99.08 \pm 0.19$ & $77.81 \pm 2.49$ & $76.67 \pm 2.58$ \\
    \midrule
    \multicolumn{6}{l}{\textit{Class model (fixed features)}} \\
    AE & $89.99 \pm 1.14$ & $93.22 \pm 0.58$ & $98.50 \pm 0.19$ & $79.70 \pm 2.15$ & $82.11 \pm 1.10$ \\
    VAE & $89.75 \pm 1.22$ & $93.44 \pm 0.55$ & $98.54 \pm 0.26$ & $69.63 \pm 2.38$ & $81.88 \pm 0.80$ \\
    PCA & $89.72 \pm 1.07$ & $95.36 \pm 0.35$ & $98.07 \pm 0.19$ & $53.83 \pm 1.18$ & $56.89 \pm 1.31$ \\
    Mean & $78.59 \pm 2.75$ & $66.54 \pm 1.98$ & $74.27 \pm 2.21$ & $23.70 \pm 0.91$ & $32.72 \pm 1.58$ \\
    Diag. Gauss. & $81.55 \pm 2.21$ & $74.00 \pm 2.18$ & $81.87 \pm 3.60$ & $26.05 \pm 1.13$ & $32.64 \pm 1.38$ \\
    \bottomrule
  \end{tabular}
\end{table}

\begin{table}[t]
  \caption{CNN pretraining coverage is evaluated by test accuracy (mean $\pm$ SD, \%, $25$ runs per cell). $N$ counts pretraining tasks. All AE uses all-class pretraining, and Linear is the corresponding CNN classifier.}
  \label{tab:pretrain_coverage}
  \centering
  \footnotesize
  \setlength{\tabcolsep}{3.0pt}
  \begin{tabular}{l*{4}{r}}
    \toprule
    $N$ / Head & \textbf{HAR} & \textbf{UWave} & \textbf{DailySp.} & \textbf{GrabMyo} \\
    \midrule
    $N{=}0$ & $89.55 \pm 0.70$ & $92.79 \pm 0.59$ & $98.69 \pm 0.21$ & $73.82 \pm 0.58$ \\
    $N{=}1$ & $90.06 \pm 1.10$ & $93.22 \pm 0.58$ & $98.52 \pm 0.26$ & $82.46 \pm 0.90$ \\
    $N{=}2$ & $92.16 \pm 1.65$ & $94.66 \pm 0.61$ & $98.81 \pm 0.19$ & $86.67 \pm 0.64$ \\
    $N{=}4$ & --- & --- & $99.15 \pm 0.26$ & $91.34 \pm 0.34$ \\
    \midrule
    All AE & $94.56 \pm 0.76$ & $96.76 \pm 0.24$ & $99.57 \pm 0.09$ & $95.19 \pm 0.20$ \\
    Linear & $94.30 \pm 0.62$ & $95.80 \pm 1.11$ & $99.09 \pm 0.24$ & $91.42 \pm 1.05$ \\
    \bottomrule
  \end{tabular}
\end{table}

\section{Conclusion}
We presented \methodname{}, a memory-efficient and replay-free framework for time series class-incremental learning. By combining frozen temporal-spectral representations with independent class-specific generative modeling, \methodname{} enables incremental learning without retaining historical data or updating previously learned models. Experiments across five TSCIL benchmarks demonstrate strong recognition performance with low memory overhead, achieving the highest average accuracy of $88.85\%$ with only $9.17$~MiB of retained state. Ablation studies further validate the effectiveness of the key design components. The modular design also ensures stable representations and predictable memory growth as new classes are introduced. These results demonstrate the potential of compact generative modeling for continual learning in resource-constrained and privacy-conscious Web-connected sensing applications.

\clearpage
\bibliographystyle{ACM-Reference-Format}
\bibliography{references}

\end{document}